\documentclass{article}

\usepackage[final,dblblindworkshop]{neurips_2026}
\workshoptitle{Continual Learning in the Era of Foundation Models and Embodied Agents}

\usepackage[utf8]{inputenc}
\usepackage[T1]{fontenc}
\usepackage{hyperref}
\usepackage{url}
\usepackage{booktabs}
\usepackage{amsmath}
\usepackage{amsfonts}
\usepackage{amssymb}
\usepackage{microtype}
\usepackage{xcolor}
\usepackage{graphicx}
\usepackage{subcaption}
\usepackage{multirow}
\usepackage{tikz}
\usepackage{placeins}

\hypersetup{hidelinks}

\title{VitaBench: Memory Lifecycles Beyond Retrieval}

\author{%
  \begin{tabular}{c}
    \textbf{Muquan Yu}\,\textsuperscript{\dag 1,2}\qquad
    \textbf{Dihan Hu}\,\textsuperscript{3} \\[0.7em]
    \textsuperscript{1}University of Hong Kong\quad
    \textsuperscript{2}Chinese University of Hong Kong\quad
    \textsuperscript{3}Cornell University \\[0.5em]
    \textsuperscript{\dag}Corresponding author: \texttt{muquan.yu@connect.hku.hk}\quad
    \texttt{dh798@cornell.edu}
  \end{tabular}
}

\begin{document}

\maketitle

\begin{abstract}
Long-term memory benchmarks for language agents primarily evaluate whether
information from earlier interactions can be retrieved and used when it becomes
relevant. Persistent memory systems, however, face an additional problem before
future use is known: deciding how a memory should persist over time---whether it
should remain exact, be compressed, consolidated, or deleted.
Existing benchmarks capture long-range retrieval, temporal reasoning, evidence
reuse, and incremental interaction, but generally do not directly evaluate the
quality of these earlier representation decisions.
We formalize this problem as \emph{memory lifecycle evaluation} and introduce
\textsc{VitaBench}, a benchmark that makes lifecycle decisions
measurable through repeated memory reviews and full-horizon counterfactual
evaluation. \textsc{VitaBench} contains a controlled Diagnostic tier that
isolates distinct future-use patterns and a Naturalistic tier that composes
them into generated customer-service-style interaction streams. We characterize lifecycle
decision structure with three diagnostics: \emph{lifecycle headroom}, which
measures the advantage of adaptive representation choices over the best fixed
strategy; \emph{horizon reversal}, which measures when short-horizon decisions
change under delayed consequences; and \emph{optimal-action diversity}, which
measures whether different memory states favor different lifecycle operations.
Across both tiers, \textsc{VitaBench} exhibits substantial lifecycle structure:
lifecycle headroom is 1.364 in Diagnostic and 1.405 in Naturalistic, and
60.2\% and 56.5\% of states, respectively, change their preferred action
between short and full horizons. A learned
delayed lifecycle policy further improves over the best fixed strategy on
\textsc{VitaBench}-Naturalistic, reaching 101.5 versus 83.4 joint utility.
These results show that retrieving past information and deciding how it should
persist are distinct evaluation problems, and that lifecycle management requires
evaluation of the delayed consequences of representation decisions.
\end{abstract}

\section{Introduction}
\label{sec:intro}

\begin{figure}[t]
\centering
\includegraphics[width=\linewidth]{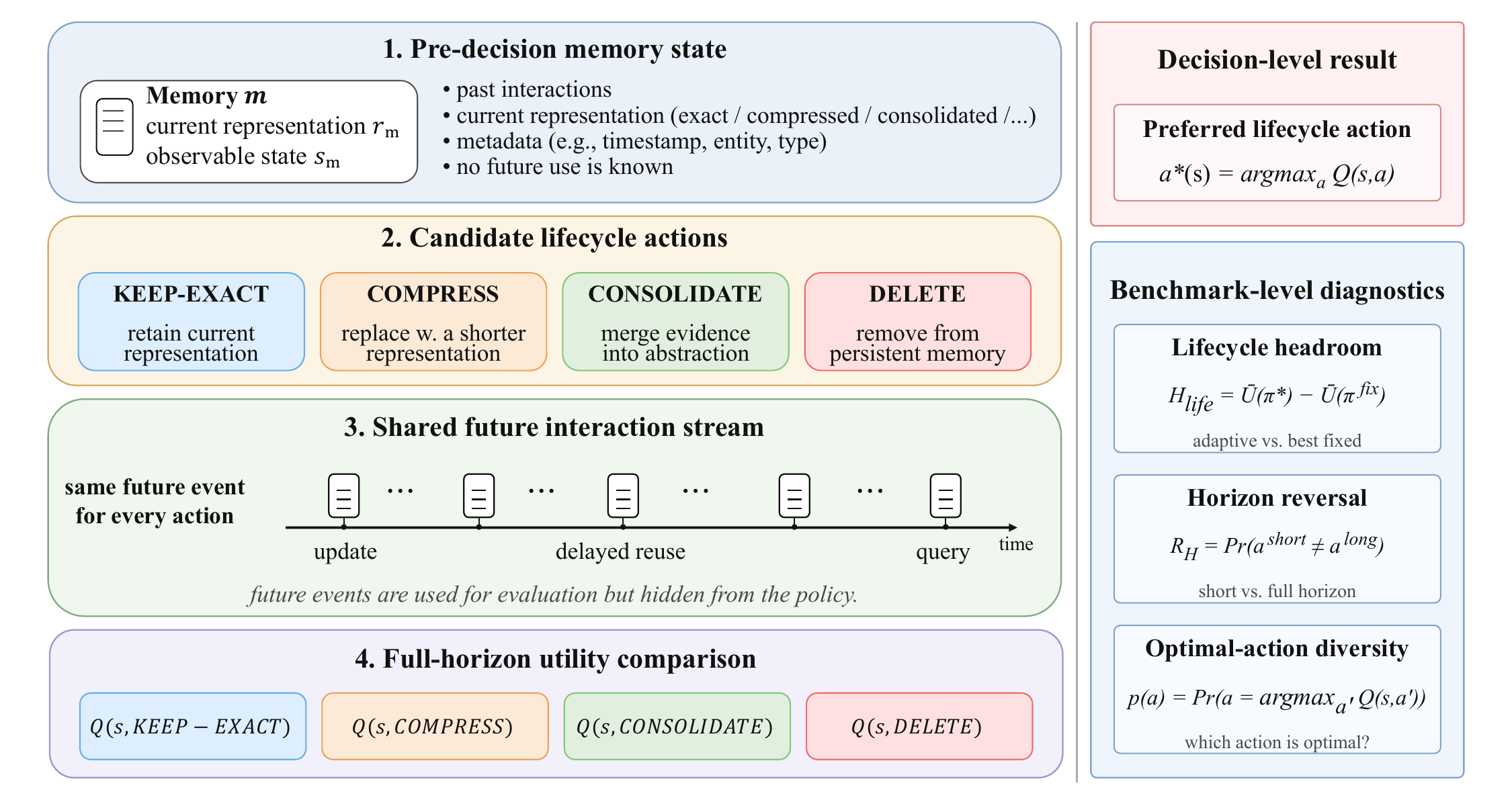}
\caption{
\textsc{VitaBench} evaluates memory lifecycle decisions. At each review point, a policy must choose how a memory should persist before its future use is known.
\textsc{VitaBench} branches from the same pre-decision memory state, applies each
candidate lifecycle action, and replays the same future interaction stream for every
branch. The resulting full-horizon utilities identify the preferred lifecycle action and
quantify the benchmark's lifecycle decision structure.
}
\label{fig:protocol}
\end{figure}

Persistent memory is becoming a central component of language agents that
operate over long interaction histories. Existing long-memory benchmarks ask
whether information observed many interactions earlier can still be recovered
and used when it becomes relevant
\citep{wu-2025-longmemeval,maharana-2024-locomo,
hu-2025-memoryagentbench,he-2026-memoryarena}. This is an important capability:
an agent that cannot recover relevant past information cannot maintain useful
long-term memory.
However, such retrieval ability is only part of the problem. A persistent memory system
must also decide what information should continue to exist and in what form,
often long before its future use is known. A memory may remain as exact
episodic evidence, be compressed into a shorter representation, be consolidated
with repeated observations, or be deleted when it appears obsolete. These
choices can have consequences much later. An exact detail that seems
unimportant today may be required by a future historical query; repeated
observations may be better represented as a compact stable preference; a
superseded fact may no longer describe the present while still being relevant
to the past. Thus, the same memory can require different representations over
its lifetime.

This creates a distinction between two evaluation problems. \emph{Retrieval evaluation} asks whether relevant
past information can be recovered when a later task arrives.
\emph{Lifecycle evaluation} asks whether the representation decisions made
\emph{before} that task arrived were appropriate. A system can succeed at the
first problem while avoiding the second---for example, by storing
everything exactly and postponing all decisions until retrieval time. Conversely,
a lifecycle policy may save storage or reduce interference by transforming
memory, but those decisions can only be judged by their downstream consequences.
Existing long-memory benchmarks were not designed to provide this supervision.
LongMemEval~\citep{wu-2025-longmemeval},
LoCoMo~\citep{maharana-2024-locomo},
MemoryAgentBench~\citep{hu-2025-memoryagentbench}, and related evaluations
such as MemoryArena~\citep{he-2026-memoryarena} contain important phenomena
including long-range retrieval, evidence reuse, temporal reasoning, updates,
and incremental interaction. Their released evaluation interfaces, however,
generally specify which past information supports a later answer rather than
what would have happened under alternative representation decisions made
earlier. Table~\ref{tab:audit} summarizes this structural distinction.

We therefore formalize \emph{memory lifecycle evaluation}: sequential
evaluation of representation choices whose value may depend on future events.
We characterize the resulting decision structure with three quantities.
\emph{Lifecycle headroom} measures how much can be gained by choosing
representations according to the current memory state rather than applying the
same fixed strategy everywhere. \emph{Horizon reversal} measures how often the
preferred action changes when short-term consequences are replaced by the full
future trajectory. \emph{Optimal-action diversity} measures whether different
memory states genuinely favor different lifecycle operations. Together, these
diagnostics answer a prerequisite question for any lifecycle benchmark:
\emph{does representation choice actually matter in this environment?}

Therefore, we introduce \textsc{VitaBench}, a benchmark designed to make these decisions
directly measurable, as shown in Figure~\ref{fig:protocol}. At each review point, a lifecycle policy observes the
current memory state and selects among \textsc{Keep-Exact},
\textsc{Compress}, \textsc{Consolidate}, and \textsc{Delete}.
\textsc{VitaBench} then branches from the same pre-decision state, applies each
candidate action, and replays the same future interaction stream. Because the
alternative futures differ only in the representation decision being evaluated,
the benchmark can compute the full-horizon consequence of each action and
identify which decisions were beneficial or harmful.
\textsc{VitaBench} contains two complementary tiers.
\textsc{VitaBench}-Diagnostic isolates four fundamental future-use patterns:
delayed exact reuse, recurrent semantic reuse, transient information, and
conflicting evidence. \textsc{VitaBench}-Naturalistic composes these phenomena
into generated customer-service-style interaction streams containing evolving entity facts,
stable preferences, obsolete information, superseding updates, historical
dependencies, and delayed reuse. The Diagnostic tier makes individual lifecycle
phenomena easy to inspect, while the Naturalistic tier tests their interaction
within a more structured multi-episode setting.
The resulting environments exhibit substantial lifecycle decision structure.
Lifecycle headroom is 1.364 utility per decision in Diagnostic and 1.405 in
Naturalistic. The preferred action changes between short and full evaluation
horizons in 60.2\% of Diagnostic states and 56.5\% of Naturalistic states.
Keeping, compressing, consolidating, and deleting
are each optimal for meaningful subsets of states, showing that lifecycle
decisions vary systematically across memory states and future-use patterns.

We further evaluate \textsc{VitaMem}, a delayed-counterfactual lifecycle policy trained to predict the long-horizon value of each lifecycle action from observable pre-decision memory features, together with fixed, heuristic, and short-horizon baselines. 
The policy is refined through policy iteration so that later lifecycle decisions are incorporated into its training targets.
On \textsc{VitaBench}-Naturalistic, repeated delayed control reaches 101.5 joint utility compared with 83.4 for the best fixed strategy. This separation shows that \textsc{VitaBench} distinguishes lifecycle policies with different temporal credit and adaptation strategies.

Our contributions are threefold. First, we formulate \emph{memory lifecycle
evaluation} as a problem distinct from long-term retrieval and introduce
lifecycle headroom, horizon reversal, and optimal-action diversity to
characterize whether an environment meaningfully exercises representation
decisions. Second, we introduce \textsc{VitaBench}, a two-tier benchmark with
repeated lifecycle decisions and full-horizon counterfactual evaluation,
enabling memory policies to be scored for decisions made before future use is
known. Third, we show that both \textsc{VitaBench} tiers contain substantial
lifecycle structure and empirically distinguish fixed, short-horizon, and
delayed lifecycle policies, providing a testbed for memory-management methods
beyond retrieval alone.

\section{Related Work}
\label{sec:related}

\begin{table}[t]
\caption{Structural comparison of representative long-memory benchmark interfaces
with respect to lifecycle-relevant capabilities. }
\label{tab:audit}
\centering
\footnotesize
\begin{tabular}{lcccc}
\toprule
Property & LongMemEval & LoCoMo & MemoryAgentBench & \textbf{\textsc{VitaBench}} (ours) \\
\midrule
Long-range memory use & yes & yes & yes & yes \\
Evidence reuse & yes & yes & yes & yes \\
Updates / conflicts & yes & partial & yes & yes \\
Incremental interaction & no & no & yes & yes \\
Repeated lifecycle decisions & no & no & no & yes \\
Counterfactual lifecycle evaluation & no & no & no & yes \\
\bottomrule
\end{tabular}
\end{table}

\paragraph{Persistent memory for language agents.}
Long-term memory systems extend language agents beyond the immediate context
through external stores, structured memories, hierarchical representations, and
evolving memory graphs. Representative systems include MemGPT, Mem0, A-MEM,
MemOS, H-MEM, HeLa-Mem, and All-Mem
\citep{packer-2023-memgpt,chhikara-2025-mem0,xu-2025-amem,
li-etal-2025-memos,sun-etal-2026-h,zhu-2026-hela,
lv-etal-2026-allmem}.

\paragraph{Learned memory management.}
Recent work treats memory maintenance as a decision problem. Pensieve,
ContextPilot, Memory-R1, Agentic Memory, and MemCon learn when or how to manage
context and persistent memory
\citep{liu-2026-statelm,pan-2026-contextpilot,yan-2026-memoryr1,
yu-etal-2026-agentic,jiang-2026-memcon}, while complementary approaches study
compression, consolidation, selective retention, stale-memory removal, and
recovery from harmful updates
\citep{dai-2026-recmem,kang-2026-retain,chen-etal-2026-dynamic-long,
kim-etal-2026-memrefine,zhang-2026-faulty,tepa-2026,
rollback-2026-repair}.

\paragraph{Long-term memory benchmarks.}
LongMemEval, LoCoMo, MemoryAgentBench, MemoryArena, LongMemEval-V2, and
LoCoMo-Plus evaluate long-range retrieval, reuse, temporal reasoning, updates,
conflicts, and incremental interaction
\citep{wu-2025-longmemeval,maharana-2024-locomo,
hu-2025-memoryagentbench,he-2026-memoryarena,
wu-2026-longmemevalv2,li-2026-locomoplus}.
These benchmarks ultimately score whether past information supports a later task.
MemOps~\citep{hao2026memopsbenchmarkinglifecyclememory} provides operation-level
supervision for memory creation, updating, deletion, and reflection, but not delayed
counterfactual evaluation of which representation action should be chosen before
future use is known.

\section{Memory Lifecycle Evaluation}
\label{sec:lifecycle}

\begin{figure}[t]
\centering
\includegraphics[width=1\linewidth]{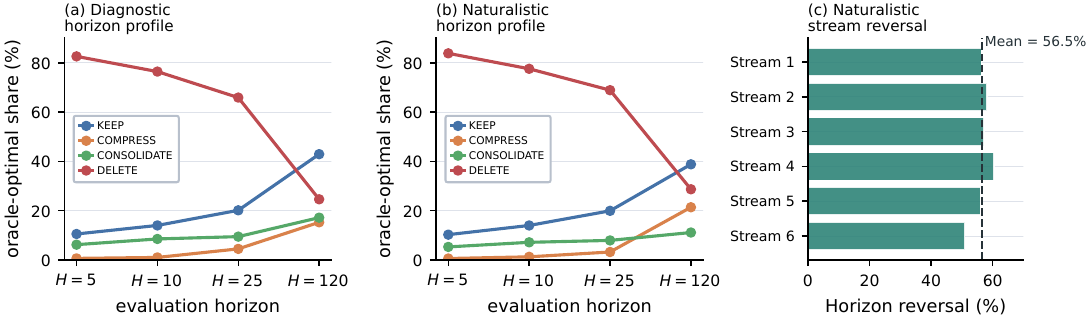}
\caption{
\textsc{VitaBench} exhibits heterogeneous and horizon-dependent lifecycle decisions.
(a) Diagnostic and (b) Naturalistic oracle-optimal action shares across \(H=5,10,25,120\);
(c) Naturalistic stream-level $H=5\rightarrow 120$ reversal rates,
with the dashed line marking the 56.5\% mean.}
\label{fig:lifecycle}
\end{figure}

We model persistent memory management as sequential choices over
memory representations.

\subsection{Lifecycle decisions}
\label{sec:lifecycle_decisions}

An agent interacts over episodes $t=1,\ldots,T$. A persistent memory $m$ has a
representation state $s_t(m)$ that records the information currently retained
about that memory together with observable metadata such as age, size, usage,
and update history. At a lifecycle review point, the policy selects an action
from
\[
\mathcal{A}
=
\{
\textsc{Keep-Exact},
\textsc{Compress},
\textsc{Consolidate},
\textsc{Delete}
\}.
\]

\textsc{Keep-Exact} preserves the current evidence without further transformation.
\textsc{Compress} replaces it with a shorter representation.
\textsc{Consolidate} combines repeated or related evidence into a more compact
semantic representation. \textsc{Delete} removes the memory from the active
persistent store.

The decision is made using only information available at the review point, but
its value can depend on events much later in the interaction. For a continuation
policy $\pi$, we define the full-horizon value of taking action $a$ in state
$s_t$ as
\[
Q^\pi(s_t,a)
=
\mathbb{E}_{\pi}
\left[
\sum_{t'=t}^{T} R_{t'}
\;\middle|\;
s_t,a
\right],
\]
where $R_{t'}$ combines downstream task reward with memory costs such as
persistent storage, retrieved context, and interference. When the continuation
policy is fixed by the evaluation protocol, we abbreviate $Q^\pi(s_t,a)$ as
$Q(s_t,a)$.

This formulation makes delayed consequences explicit: compression can reduce
immediate cost while losing a later exact detail, consolidation can improve
semantic reuse, and deletion can be beneficial unless obsolete information is
unexpectedly needed again.

\subsection{When does an environment exercise lifecycle management?}
\label{sec:lifecycle_structure}

A long interaction history does not by itself imply a meaningful lifecycle
problem. If the same representation action is optimal for nearly every state,
then a fixed strategy is sufficient even when retrieval remains difficult. We
therefore characterize lifecycle decision structure with three benchmark-level
diagnostics.

\paragraph{Lifecycle headroom.}
Let $\bar{U}(\pi)$ denote the expected joint utility of lifecycle policy $\pi$,
normalized per lifecycle decision. Let $\pi^\star$ denote the oracle lifecycle policy that selects the
full-horizon optimal action at each decision point, and let $\pi^{\mathrm{fix}}$ denote the best policy
that applies one fixed lifecycle action throughout the environment. We define
\[
H_{\mathrm{life}} = \bar{U}(\pi^\star) - \bar{U}(\pi^{\mathrm{fix}}).
\]
Larger headroom indicates greater potential benefit from adapting the representation
to the current memory state at each review point.

\paragraph{Horizon reversal.}
Lifecycle actions can have different rankings depending on how far into the
future they are evaluated. Let
$a^\star_H(s)$ denote the action with highest utility when consequences are
evaluated over horizon $H$. For a short horizon $H_s$ and a longer horizon
$H_\ell$, we define
\[
R_H
=
\Pr_s
\left[
a^\star_{H_s}(s)
\neq
a^\star_{H_\ell}(s)
\right].
\]
A high reversal rate indicates that short-term evaluation frequently gives a
different decision from full-horizon evaluation, making delayed credit
assignment important.

\paragraph{Optimal-action diversity.}
Finally, we utilize optimal-action diversity to measure how oracle-optimal actions are distributed across decision
states:
\[
p(a)
=
\Pr_s
\left[
a
=
\arg\max_{a' \in \mathcal{A}} Q(s,a')
\right].
\]
If one action dominates almost all states, the benchmark offers little reason
for state-dependent lifecycle control. A heterogeneous distribution instead
indicates that different memory states and future-use patterns favor different
representation choices.

Together, lifecycle headroom, horizon reversal, and optimal-action diversity
describe complementary aspects of the benchmark: whether adaptive decisions
can improve utility, whether those decisions depend on delayed consequences,
and whether multiple lifecycle operations are genuinely useful.

\begin{table}[t]
\caption{\textsc{VitaBench} structural diagnostics on held-out streams. Action shares
are fractions of states for which each action is full-horizon oracle-optimal.}
\label{tab:structure}
\centering
\footnotesize
\begin{tabular}{lcccccc}
\toprule
Tier & $H_{\rm life}$ & $R_H$ & KEEP & COMPRESS & CONSOLIDATE & DELETE \\
\midrule
Diagnostic & 1.364 & 0.602 & 0.429 & 0.153 & 0.172 & 0.246 \\
Naturalistic & 1.405 & 0.565 & 0.388 & 0.214 & 0.111 & 0.287 \\
\bottomrule
\end{tabular}
\end{table}

\subsection{Evaluating lifecycle policies}
\label{sec:lifecycle_policy_metrics}

Once an environment is characterized, individual lifecycle policies are evaluated
by the consequences of the actions they select.

\paragraph{Joint utility.}
We evaluate the overall task--memory trade-off using joint utility,
\[
U(\pi)
=
\sum_{t=1}^{T}
\left(
r_t^{\mathrm{task}}
-\lambda_{\mathrm{active}} c_t^{\mathrm{active}}
-\lambda_{\mathrm{storage}} c_t^{\mathrm{storage}}
-\lambda_{\mathrm{retrieval}} c_t^{\mathrm{retrieval}}
-\lambda_{\mathrm{interference}} c_t^{\mathrm{interference}}
\right),
\]
where $r_t^{\mathrm{task}}$ is the downstream task reward at episode $t$, and the
$\lambda$ terms weight the costs of active memory, persistent storage, retrieved
context, and interference, respectively. Their precise definitions and coefficients
are reported in Appendix~\ref{app:operators}.

\paragraph{Persistent storage.}
We separately report the mean size of the maintained persistent memory store,
\[
S(\pi)
=
\frac{1}{T}\sum_{t=1}^{T} |\mathcal{M}_t^\pi|,
\]
measured in persistent tokens per episode. Reporting storage separately makes the
memory footprint directly interpretable even though storage cost also contributes
to joint utility.

\paragraph{Policy regret.}
For the common held-out decision states $s_1,\ldots,s_N$ with full-horizon
counterfactual values $Q_i(a)$, the regret of policy $\pi$ is
\[
\operatorname{Regret}(\pi)
=
\frac{1}{N}\sum_{i=1}^{N}
\left[
\max_{a\in\mathcal{A}} Q_i(a)
-
Q_i\!\left(\pi(s_i)\right)
\right].
\]
Regret measures how much full-horizon utility is lost by the actions selected by a
policy relative to the best action available at each state. The implementation
averages this quantity over the same held-out review-point set for every policy.

The three structural diagnostics characterize the lifecycle decision problem itself,
while joint utility, storage, and regret characterize the policies evaluated on that
problem.

\section{\textsc{VitaBench}}
\label{sec:vitabench}

\begin{table}[tb]
\caption{Policy evaluation on held-out \textsc{VitaBench} tiers.
Entries report joint utility, mean persistent storage in tokens per episode,
and full-horizon policy regret.}
\label{tab:policy_main}
\centering
\footnotesize
\begin{tabular}{lrrr|rrr}
\toprule
& \multicolumn{3}{c}{Diagnostic} & \multicolumn{3}{c}{Naturalistic} \\
Policy & Joint utility $\uparrow$ & Storage $\downarrow$ & Regret $\downarrow$ & Joint utility $\uparrow$ & Storage $\downarrow$ & Regret $\downarrow$ \\
\midrule
always \textsc{Keep-Exact} & 52.4 & 9{,}722 & 1.50 & 29.3 & 7{,}028 & 1.41 \\
always \textsc{Compress} & 170.1 & \textbf{629} & 1.40 & 83.4 & \textbf{413} & 2.53 \\
always \textsc{Consolidate} & 125.4 & 4{,}418 & 1.50 & 40.5 & 2{,}776 & 2.61 \\
myopic ($H=1$) & 55.5 & 878 & 1.90 & 10.6 & 613 & 2.78 \\
one-shot delayed & 171.4 & 3{,}026 & 1.50 & 64.1 & 5{,}668 & 1.32 \\
repeated \textsc{VitaMem} & \textbf{200.4} & 1{,}539 & \textbf{1.10} & \textbf{101.5} & 2{,}616 & \textbf{0.86} \\
oracle & 231.0 & 2{,}718 & 0.00 & 142.6 & 2{,}468 & 0.00 \\
\bottomrule
\end{tabular}
\end{table}

\textsc{VitaBench} evaluates how persistent memories are represented over time
and is organized into two complementary tiers: \textsc{VitaBench}-Diagnostic
and \textsc{VitaBench}-Naturalistic. At each lifecycle review point, a policy
observes the current memory state and selects one of four representation actions:
\textsc{Keep-Exact}, \textsc{Compress}, \textsc{Consolidate}, or
\textsc{Delete}. The benchmark specifies the interaction stream, memory
operators, downstream tasks, and memory costs, so that policies are compared
under the same environment and action semantics.

The central feature of \textsc{VitaBench} is counterfactual lifecycle
evaluation. For a given review state, the benchmark independently applies each
valid action and then replays the same subsequent interaction stream. The
resulting trajectories differ only in the representation decision made at that
review point, allowing the full-horizon utility of alternative actions to be
measured directly. 
As illustrated in
Figure~\ref{fig:protocol}, the resulting full-horizon utilities identify the
preferred action and support the structural diagnostics defined in
Section~\ref{sec:lifecycle}.

\subsection{\textsc{VitaBench}-Diagnostic}
\label{sec:vitabench_diagnostic}

The \textsc{VitaBench}-Diagnostic tier isolates four recurring lifecycle patterns in controlled
interaction streams. Each stream contains 60 persistent memory threads evolving
over 120 episodes. \emph{Exact} memories contain details that may be required
after a long delay, favoring preservation when precise information remains
important. \emph{Recurrent} memories are reused repeatedly and can benefit from
more compact semantic representations. \emph{Transient} memories often become
irrelevant but may occasionally be reused, creating a tradeoff between deletion
and future availability. \emph{Conflict} memories receive incompatible updates
whose later resolution makes premature consolidation costly. These processes
account for 30\%, 25\%, 25\%, and 20\% of memory threads, respectively.

At each review point, the selected representation determines subsequent
task performance and memory cost. The evaluator accounts for persistent
storage, retrieved context, active-memory cost, and interference while scoring
future queries from the same underlying stream. Because every candidate
action can be evaluated from the same state, the Diagnostic tier provides
full-horizon action values computed exactly under the specified operator model,
from which lifecycle headroom, horizon reversal, and optimal-action diversity
are computed, and against which policies are scored.

The tier isolates delayed exact reuse, recurrent semantic reuse, transient
information, and conflicting evidence, making the sources of lifecycle
difficulty directly inspectable.

\subsection{\textsc{VitaBench}-Naturalistic}
\label{sec:vitabench_naturalistic}

The \textsc{VitaBench}-Naturalistic tier composes these lifecycle phenomena into
generated customer-service-style interaction streams. Each stream contains 40 persistent
entities evolving over 120 episodes. Memories correspond to entity
attributes, preferences, transactions, and updates, while later requests may
depend on current facts, earlier historical states, or multiple memories
jointly.

The streams combine delayed exact reuse, recurring preferences, temporary
information, obsolescence, superseding updates, historical queries, and
conflicting evidence. A later request may refer to an earlier transaction while
repeated preferences favor consolidation and one-time instructions can safely
disappear. Thus preservation, abstraction, and removal have different
consequences depending on memory history.

Naturalistic uses the same lifecycle actions and counterfactual evaluation
principle as Diagnostic. The distinction between the two tiers is 
one of task structure rather than evaluation semantics: Diagnostic isolates
individual future-use patterns, whereas Naturalistic combines them within
multi-episode entity histories and composite tasks. Both tiers use disjoint training and held-out evaluation streams; generator parameters, operator fidelities, costs, and random seeds are provided in
Appendix~\ref{app:env}.

\section{A Reference Lifecycle Policy: \textsc{VitaMem}}
\label{sec:method}

\begin{figure}[tb]
\centering
\includegraphics[width=1\linewidth]{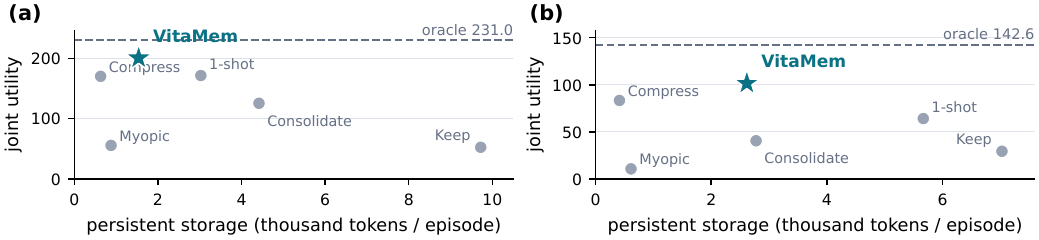}
\caption{
Repeated \textsc{VitaMem} improves the storage--utility trade-off over
fixed and myopic policies.
Results are shown on held-out \textsc{VitaBench} tiers:
(a) Diagnostic and (b) Naturalistic. The star marks repeated \textsc{VitaMem},
and dashed lines indicate oracle joint utility.
}
\label{fig:storage_utility}
\end{figure}

We evaluate \textsc{VitaMem}, a lifecycle policy trained from the
counterfactual outcomes provided by \textsc{VitaBench}. At each review point,
the policy observes only information available at that time, including the
memory's age, size, usage statistics, exact-detail indicators, update or
conflict state, and current memory costs. It predicts the relative value of
the valid lifecycle actions and selects the highest-valued action.

For a continuation policy $\pi$, the training target for taking action $a$ in
state $s_t$ is chosen to be the full-horizon value introduced in Section~\ref{sec:lifecycle},
\[
Q^\pi(s_t,a)
=
\mathbb{E}_{\pi}
\left[
\sum_{t'=t}^{T} R_{t'}
\mid s_t,a
\right].
\]

\textsc{VitaBench} constructs these targets by applying each candidate action
to the same pre-decision state and replaying the same future interaction
stream. 
Future events enter the offline counterfactual target construction but are never
available to the policy at decision time, and \textsc{VitaMem} receives only
pre-decision features.
We fit a two-hidden-layer MLP with 48 and 24 hidden units to predict the relative full-horizon values of the valid lifecycle actions.

Because the value of one representation decision depends on decisions made
later in the same memory lifecycle, we refine the targets through policy
iteration. At iteration $k$, each candidate action is evaluated at the current
review point while subsequent reviews follow the current policy $\pi_k$.
The resulting counterfactual values are then used to train the next policy.
We evaluate both \emph{one-shot} control, where each memory is transformed at
most once, and \emph{repeated} control, where the same memory can be
reconsidered as its history evolves. Additional credit-assignment and training
details are provided in Appendices~\ref{app:credit}--\ref{app:joint}.

\section{Experiments}
\label{sec:experiments}

We evaluate \textsc{VitaBench} in three steps. We first ask whether its two
tiers exhibit the lifecycle decision structure defined in
Section~\ref{sec:lifecycle_structure}. We then test whether this structure
produces meaningful separation among lifecycle policies. Finally, we compare
the evaluation interface of \textsc{VitaBench} with representative long-memory
benchmarks.

\subsection{Does \textsc{VitaBench} expose lifecycle decision structure?}
\label{sec:exp_structure}

Table~\ref{tab:structure} reports lifecycle headroom, horizon reversal, and the
distribution of full-horizon optimal actions. For horizon
reversal, we compare the preferred action at $H=5$ with that under the full
$H=120$ trajectory.
Both tiers exhibit substantial lifecycle headroom: adapting representation
choices to individual memory states improves over the strongest fixed action by
1.364 utility per decision in Diagnostic and 1.405 in Naturalistic. The
preferred action also depends strongly on delayed consequences. 
In 60.2\% of Diagnostic states and 56.5\% of Naturalistic states, the action
preferred at $H=5$ differs from the action preferred over the full $H=120$
trajectory.

Optimal actions are likewise heterogeneous. In Diagnostic,
\textsc{Keep-Exact}, \textsc{Compress}, \textsc{Consolidate}, and
\textsc{Delete} are preferred in 42.9\%, 15.3\%, 17.2\%, and 24.6\% of
states, respectively. Naturalistic exhibits a similarly broad distribution:
38.8\%, 21.4\%, 11.1\%, and 28.7\%. Thus, lifecycle decisions vary both across
memory states and across evaluation horizons.

Figure~\ref{fig:lifecycle} further shows that lifecycle-optimal decisions depend strongly on evaluation horizon.
Action preferences shift substantially as delayed consequences become visible, with $60.2\%$ of Diagnostic states and $56.5\%$ of Naturalistic states reversing between $H=5$ and $H=120$, while multiple lifecycle operations remain useful across states.
These results indicate that lifecycle decisions should be evaluated beyond their immediate consequences.

\subsection{Can \textsc{VitaBench} distinguish lifecycle policies?}
\label{sec:exp_policy}

We next evaluate policies that differ in how they use state information and
future credit. All methods operate on identical interaction streams with the
same lifecycle actions, memory operators, costs, and evaluator. We compare
three fixed representations, myopic control, one-shot delayed
\textsc{VitaMem}, repeated delayed \textsc{VitaMem}, and the full-horizon
oracle.

Fixed representations already differ substantially: joint utility ranges from
52.4 to 170.1 on Diagnostic and from 29.3 to 83.4 on Naturalistic.
Repeated delayed
\textsc{VitaMem} improves over the strongest fixed strategy, reaching 200.4
versus 170.1 on Diagnostic and 101.5 versus 83.4 on Naturalistic. The
Naturalistic gain is $+18.1$ (paired 95\% bootstrap interval $[5.7,31.2]$).
Figure~\ref{fig:storage_utility} shows the corresponding storage--utility
trade-off: repeated \textsc{VitaMem} lies above the fixed-policy points while
retaining substantially less storage than always \textsc{Keep-Exact}.
Myopic control reaches only 55.5 on Diagnostic and 10.6 on Naturalistic;
repeated control reaches 101.5 versus 64.1 for one-shot delayed control on
Naturalistic. The remaining oracle gaps are 30.6 on Diagnostic and 41.1 on
Naturalistic.

\subsection{Relationship to existing long-memory benchmarks}
\label{sec:exp_audit}

We compare the lifecycle-evaluation interfaces of \textsc{VitaBench},
LongMemEval~\citep{wu-2025-longmemeval}, LoCoMo~\citep{maharana-2024-locomo},
and MemoryAgentBench~\citep{hu-2025-memoryagentbench} in Table~\ref{tab:audit}.
These benchmarks support retrieval, reuse, and updates, but their released
evaluation interfaces do not specify counterfactual outcomes for earlier
representation choices.

\FloatBarrier

\section{Discussion}
\label{sec:discussion}

\textsc{VitaBench} separates retrieval from lifecycle management by evaluating
representation decisions before future use is known. Both the Diagnostic and
Naturalistic tiers exhibit lifecycle headroom, horizon reversal, and optimal-action
diversity, while the separation among fixed, myopic, and delayed policies shows that
these structural properties translate into meaningful differences in lifecycle
control. These results establish memory lifecycle management as a distinct evaluation axis
for persistent-memory language agents alongside long-term retrieval.

\newpage
{
\small
\bibliographystyle{plainnat}
\bibliography{references_v3}
}

\newpage
\appendix
\appendix
\section*{Appendices}

\section{Benchmark Generation and Environment Details}
\label{app:env}

Both \textsc{VitaBench} tiers use the lifecycle action space and
counterfactual evaluation protocol described in
Sections~\ref{sec:lifecycle} and~\ref{sec:vitabench}.
\textsc{VitaBench} is designed as a controlled evaluation environment:
its purpose is to isolate the consequences of representation decisions while
holding the future interaction stream fixed, rather than to reproduce the
behavior of any particular deployed memory system. Generator parameters and
lifecycle operators therefore define interpretable representation trade-offs
under which counterfactual action values can be evaluated exactly.
Appendix~\ref{app:realtext} separately tests whether the main qualitative
assumptions underlying these controlled operators are consistent with
transformations observed on natural long-memory text.

This appendix specifies the generation process, operator semantics, utility
terms, and disjoint design, training, and evaluation splits used in the
experiments.

\subsection{\textsc{VitaBench}-Diagnostic}

Each Diagnostic stream contains 60 persistent memory threads evolving over
120 episodes. Threads are assigned to four future-use processes: Exact (30\%),
Recurrent (25\%), Transient (25\%), and Conflict (20\%).

Exact memories contain information whose precise value is queried 10--66
episodes after creation. Recurrent memories contain 3--5 related observations
and 3--5 future uses; 25\% of these queries require exact detail, while the
remainder depend on recurring semantic content. Transient memories are usually
obsolete after their initial role but have a 12\% probability of unexpected
future reuse. Conflict memories contain contradictory observations and are
resolved later with probability 30\%.

Training uses eight streams with seeds 0--7. Final evaluation uses six
disjoint streams with seeds 1000--1005.

\subsection{\textsc{VitaBench}-Naturalistic}

The Naturalistic tier contains customer-service-style histories for 40
persistent entities evolving over 120 episodes. Each entity accumulates facts,
preferences, temporary instructions, updates, and conflicting evidence that
may later be reused by downstream tasks.

The generator mixes Exact, Recurrent, Transient, and Conflict future-use
processes with probabilities 0.20, 0.25, 0.30, and 0.25, respectively. Exact
details are reused after delays of 24--60 episodes. Recurrent memories model
stable preferences and repeated attributes with both semantic and exact reuse.
Transient memories represent one-time or obsolete information with occasional
surprise reuse, while Conflict memories contain incompatible updates that are
resolved with probability 0.45.

These processes are composed into entity histories containing delayed exact
reuse, stable preferences, temporary exceptions, superseding updates,
historical queries, and conflicting evidence. Queries may depend on the
current value of an attribute, an earlier historical value, or several
memories jointly.

The Naturalistic generator was finalized using separate design streams before
any learned-policy evaluation. We considered six interpretable configurations
that varied the prevalence and timing of the four future-use processes.
Lifecycle headroom, horizon reversal, and optimal-action diversity were used
as design diagnostics to avoid configurations in which representation choice
was effectively degenerate; learned-policy performance was not used for
configuration selection.

All subsequent policy training and final evaluation use disjoint streams.
Training uses seeds 0--7, final evaluation uses seeds 6000--6005, and the
design sweep uses separate seeds 5000 and 5001. Robustness across alternative
generator regimes is reported in Appendix~\ref{app:sensitivity}.

\subsection{Lifecycle operators and utility}
\label{app:operators}

All policies within a tier use the same lifecycle operators. These operators
are intentionally abstract and controlled: their fidelity parameters are not
intended as universal empirical estimates of LLM compression or consolidation
quality. Instead, they encode qualitatively distinct representation
trade-offs needed to evaluate lifecycle decisions. Exact retention preserves
detail at higher storage cost, compression trades detail for compactness,
consolidation favors recurring semantic information over instance-specific
detail, and deletion removes both retained information and future storage
cost. The same operator definitions are used for every compared policy.

The canonical parameters define a fixed evaluation environment.
\textsc{Keep-Exact} retains the current representation without information
loss. \textsc{Compress} reduces the representation to 35\% of its original
token count, with success probability 0.40 for exact-detail queries and 0.70
for semantic-rule queries. \textsc{Consolidate} combines $k$ related
observations into a representation of size $40+5k$ tokens, succeeds with
probability 0.90 on rule-level queries and 0.05 on exact-detail queries, and
can be harmful when unresolved conflicting evidence is merged.
\textsc{Delete} removes the active representation and incurs no subsequent
storage cost, but cannot answer queries that require the deleted information.

These canonical values are held fixed across the main experiments rather than
fit to the evaluated policies. Appendix~\ref{app:realtext} provides an
independent construct check using transformations measured on natural
long-memory text, while Appendix~\ref{app:sensitivity} examines the stability
of the benchmark's structural conclusions across alternative generator
regimes.

Task rewards and memory costs are fixed before policy evaluation and shared by
all methods. The canonical setting assigns reward 8 to exact-detail queries
and reward 2 to semantic-rule queries. Together with the memory-cost terms,
these fixed weights define the joint-utility trade-off used in the main
experiments:
\[
\lambda_{\mathrm{active}}=0.002,\qquad
\lambda_{\mathrm{storage}}=0.00025,\qquad
\lambda_{\mathrm{retrieval}}=0.01,\qquad
\lambda_{\mathrm{interference}}=0.5.
\]
These coefficients define the canonical balanced-cost regime and should be
interpreted as benchmark trade-off parameters rather than
deployment-specific economic costs.

\subsection{Counterfactual evaluation}

At each review point, the evaluator saves the complete pre-decision state.
Each valid lifecycle action is then applied independently to that same state,
after which every branch is replayed through the identical subsequent
interaction stream. Consequently, the branches differ only in the
representation decision under evaluation rather than in future events.

The full-horizon oracle action is
\[
a^\star(s_t)=\arg\max_{a\in\mathcal{A}} Q(s_t,a),
\]
where $Q(s_t,a)$ accumulates utility over the remainder of the trajectory.
Short-horizon values use the same starting state and future stream truncated
to the specified horizon; the main horizon-reversal analysis compares $H=5$
with the full $H=120$ trajectory.

Because the lifecycle operators have explicit probabilistic semantics, the
evaluator can compute expected query utility analytically while replaying the
shared future stream. Thus, ``exact counterfactual evaluation'' refers to
exact expected utility under the specified benchmark model, not to a claim
that the operator model itself exactly reproduces real-world memory
transformations. The reported action values are therefore free of Monte Carlo
variation from sampled operator outcomes.

\section{Full Policy Results}
\label{app:policy_full}

The main text reports a two-tier comparison of fixed, myopic, delayed, and
oracle policies. Tables~\ref{tab:diag_full} and~\ref{tab:nat_full} provide the
full held-out results. All policies within each tier use the same interaction
streams, lifecycle operators, costs, and evaluator.

\begin{table}[h]
\caption{Full held-out policy evaluation on
\textsc{VitaBench}-Diagnostic. Storage is mean persistent tokens per episode;
regret is the mean full-horizon oracle-minus-selected-action value per review
point over the common held-out decision-point set.}
\label{tab:diag_full}
\centering
\footnotesize
\begin{tabular}{lrrr}
\toprule
Policy & Joint utility $\uparrow$ & Storage $\downarrow$ & Regret $\downarrow$ \\
\midrule
recency delete & $-57.6$ & 2{,}986 & 2.5 \\
always \textsc{Keep-Exact} & 52.4 & 9{,}722 & 1.5 \\
myopic ($H=1$) & 55.5 & 878 & 1.9 \\
always \textsc{Consolidate} & 125.4 & 4{,}418 & 1.5 \\
tuned exact-aware heuristic & 130.3 & 6{,}506 & \textbf{1.1} \\
always \textsc{Compress} & 170.1 & \textbf{629} & 1.4 \\
one-shot delayed & 171.4 & 3{,}026 & 1.5 \\
repeated \textsc{VitaMem} & \textbf{200.4} & 1{,}539 & \textbf{1.1} \\
oracle & 231.0 & 2{,}718 & 0.0 \\
\bottomrule
\end{tabular}
\end{table}

Repeated \textsc{VitaMem} improves over always-\textsc{Compress}, the
strongest fixed Diagnostic policy, by $+30.3$ joint utility with a paired
95\% interval of $[18.0,48.2]$ across the six held-out streams.

\begin{table}[h]
\caption{Full held-out policy evaluation on
\textsc{VitaBench}-Naturalistic over six evaluation streams. Storage is mean
persistent tokens per episode; regret is averaged over the common held-out
decision-point set.}
\label{tab:nat_full}
\centering
\footnotesize
\begin{tabular}{lrrr}
\toprule
Policy & Joint utility $\uparrow$ & Storage $\downarrow$ & Regret $\downarrow$ \\
\midrule
always \textsc{Keep-Exact} & 29.3 & 7{,}028 & 1.41 \\
always \textsc{Compress} & 83.4 & 413 & 2.53 \\
always \textsc{Consolidate} & 40.5 & 2{,}776 & 2.61 \\
always \textsc{Delete} & $-2.6$ & \textbf{88} & 4.42 \\
recency heuristic & $-5.3$ & 219 & 3.36 \\
tuned heuristic & 78.0 & 4{,}782 & 1.24 \\
myopic ($H=1$) & 10.6 & 613 & 2.78 \\
one-shot delayed & 64.1 & 5{,}668 & 1.32 \\
repeated \textsc{VitaMem} & \textbf{101.5} & 2{,}616 & \textbf{0.86} \\
oracle & 142.6 & 2{,}468 & 0.00 \\
\bottomrule
\end{tabular}
\end{table}

Repeated \textsc{VitaMem} exceeds the strongest fixed Naturalistic policy by
$+18.1$ joint utility, with a paired 95\% bootstrap interval of
$[5.7,31.2]$. Relative to the myopic controller, the improvement is
$+90.9$ $[73.0,111.1]$. The oracle exceeds repeated control by
$41.1$ $[31.1,51.4]$.

\section{Delayed Credit Assignment}
\label{app:credit}

We test whether lifecycle learning requires exhaustive counterfactual
branching or primarily requires sufficiently long credit assignment. Using the
same Q-advantage architecture, we compare full counterfactual targets with two
reduced-branching variants and immediate next-episode supervision.

This is the delayed-credit ablation configuration; its utilities are separate
from the canonical joint-aware repeated-policy result reported in
Table~\ref{tab:policy_main}.

Full counterfactual supervision obtains 208.2 held-out joint utility. MC10,
which branches over all decision points using ten sampled continuations,
obtains 209.2. MC3, which branches over the 50\% most sensitive decision
points using three continuations, obtains 216.3. In contrast, supervision
restricted to the next episode ($H=1$) reaches only 55.5.

\begin{table}[h]
\caption{Delayed-credit ablation configuration on held-out Diagnostic streams;
these values are separate from the canonical joint-aware repeated-policy
result in Table~\ref{tab:policy_main}.}
\label{tab:credit_ablation}
\centering
\footnotesize
\begin{tabular}{lr}
\toprule
Training target & Joint utility $\uparrow$ \\
\midrule
full-horizon counterfactual & 208.2 \\
MC10 reduced branching & 209.2 \\
MC3 sensitive-state branching & 216.3 \\
immediate ($H=1$) & 55.5 \\
\bottomrule
\end{tabular}
\end{table}

The limited-branch variants are comparable to full counterfactual supervision
at the stream level, whereas immediate supervision fails sharply. The primary
requirement is therefore long-horizon credit rather than exhaustive branching
of every future.

\section{Joint-Aware Counterfactual Credit}
\label{app:joint}

The value of a lifecycle action depends on the decisions made at later review
points. Counterfactual labels generated against a fixed
\textsc{Keep-Exact}-everywhere continuation can therefore overvalue destructive
actions: after one memory is compressed or deleted, later decisions made by the
learned policy may change as well.

We address this dependence through policy iteration. At iteration $k$, states
are generated under the current policy $\pi_k$. At each review point, one
candidate action is substituted while both the factual and counterfactual
branches subsequently continue under $\pi_k$. These joint-aware targets are
then used to obtain the next policy.

After three iterations, joint-aware training reaches the canonical paired
held-out mean of 200.4 across six streams, compared with 170.1 for
always-\textsc{Compress}. The paired difference is $+30.3$
$[18.0,48.2]$ and is positive on all six held-out streams. In a
cost-sensitive regime, joint-aware credit reaches 112.0, compared with 65.5
for single-deviation training and 108.1 for always-\textsc{Compress}.

\section{Real-Text Construct Checks}
\label{app:realtext}

The canonical benchmark uses controlled operators so that alternative
lifecycle actions can be evaluated exactly. We additionally test two
assumptions underlying these operators on natural long-memory text.

\paragraph{Representation transformations are lossy.}
Using Qwen2.5-0.5B-Instruct on LongMemEval sessions,
\textsc{Compress} retains 0.21 of identifiable exact-value tokens and
\textsc{Consolidate} retains 0.27. A consolidation prompt that explicitly
preserves both sides of conflicting evidence reduces measured
overgeneralization to 8\%.

Replacing the controlled operator fidelities with these empirically measured
values preserves the policy separation on the six held-out Diagnostic streams
in the real-text-calibrated operator regime: \textsc{VitaMem} reaches 213.0 joint
utility compared with 111.6 for always-\textsc{Compress}, a paired difference
of $+101.4$ $[83.4,122.6]$ across six streams. In the cost-sensitive setting,
the corresponding difference is $+4.7$ $[0.9,8.7]$.

\paragraph{Useful evidence is not determined by recency alone.}
On LongMemEval, retaining the newest 16 sessions preserves 0.392 of answer
evidence, compared with 0.828 for a learned lexical selector and 0.937 for a
semantic selector. On LoCoMo, leave-one-conversation-out semantic selection
reaches 0.723 evidence recall, exceeding keep-newest-16 by $+0.115$
$[0.088,0.142]$.

\section{Rollback and Version Preservation}
\label{app:rollback}

The canonical benchmark uses the four actions
\textsc{Keep-Exact}, \textsc{Compress}, \textsc{Consolidate}, and
\textsc{Delete}. We separately evaluate an extension in which earlier
representations are archived and can be restored through \textsc{Rollback}.

A learned controller with reversible representations and \textsc{Rollback}
reaches 185.8 joint utility, compared with 159.2 for the matched
rollback-extension controller without rollback, an improvement of $+26.6$
$[21.0,34.4]$. The average number
of harmful consolidation actions falls from 14.8 to 4.5 per stream.

A fixed rollback-on-resolution heuristic reaches only 100.2, below the
destructive consolidation baseline of 125.4. Version preservation is therefore
most useful when restoration is selected conditionally rather than applied as
a fixed rule.

\section{Low-Headroom Customer-Service Control}
\label{app:low_headroom}

We evaluate an alternative customer-service generator as a negative control for
the lifecycle-structure diagnostics. It uses the same 40-entity, 120-episode
setting and lifecycle action interface as the Naturalistic tier, but contains
substantially less opportunity for representation-dependent decisions.

At balanced costs, its full-horizon oracle-optimal actions are distributed as
17\% \textsc{Keep-Exact}, 23\% \textsc{Compress}, 13\%
\textsc{Consolidate}, and 48\% \textsc{Delete}. Its lifecycle headroom is
0.49 utility per decision, substantially below the 1.405 measured for the
Naturalistic benchmark used in the main experiments, and its delayed deletion
reversal rate is 19.4\%.

This control shows that exposing the same lifecycle action space does not by
itself produce strong lifecycle decision structure. The proposed diagnostics
therefore distinguish environments in which adaptive representation choice has
substantial potential value from those in which a largely fixed strategy is 
sufficient.

\section{Policy Behavior Across Memory Regimes}
\label{app:regimes}

The oracle preferences in Diagnostic vary systematically with the future-use
process. Exact memories place high value on preserving precise information
because later queries may require details that lossy representations cannot
recover. Recurrent memories more often favor compression or consolidation,
which reduce repeated storage while preserving semantic information. Transient
memories frequently favor compression or deletion, while Conflict memories
penalize irreversible consolidation when competing historical evidence remains
relevant.

These regime-specific preferences explain why no single representation action
dominates the benchmark. They also connect the aggregate action-diversity
statistics to distinct forms of future memory use rather than to random
variation across states.

\section{Public-Benchmark Audit and Construct-Check Protocol}
\label{app:audit}

The structural comparison in Section~\ref{sec:exp_audit} uses
LongMemEval-S~\citep{wu-2025-longmemeval},
LoCoMo~\citep{maharana-2024-locomo}, and
MemoryAgentBench~\citep{hu-2025-memoryagentbench}.
LongMemEval-S contains 500 instances with approximately 48 haystack sessions
per instance; 74\% of annotated evidence occurs more than ten session positions
before the end of the haystack, while knowledge-update and temporal-reasoning
questions account for 15.6\% and 26.6\% of instances, respectively.
LoCoMo contains substantial evidence reuse: 78\% of evidence references support
multiple questions, and 21\% of questions require multiple evidence passages.
MemoryAgentBench presents experience incrementally and includes explicit
conflict-resolution tasks. These properties make the benchmarks suitable for
evaluating long-range and continually reused memory, but their released
interfaces do not directly assign counterfactual values to alternative
representation decisions made earlier in the interaction.

For the real-text construct checks in Appendix~\ref{app:realtext}, we use a
question-instance-level 80/20 split of LongMemEval-S with seed 0, producing
400 training and 100 test question IDs. Test questions, answers, and evidence
labels are excluded from training. Because LongMemEval draws haystack sessions
from a shared conversation pool, some session overlap across instances is
unavoidable; only 16 of 948 evidence sessions are shared across the train/test
split. A minimized-overlap rerun yields semantic and lexical evidence recall of
0.901 and 0.800, respectively, compared with 0.381 for keep-newest-16.
LoCoMo selection experiments use leave-one-conversation-out evaluation over its
ten conversations. In all construct checks, deployment-time features exclude
question answers, gold evidence labels, and future-event information.

\section{More Detailed Discussion on the Relationship to Existing Memory Benchmarks}
\label{app:benchmark_comparison}

\textsc{VitaBench} is intended to complement, rather than replace, existing
long-term memory benchmarks. The main distinction is the \emph{object being
evaluated}. Existing benchmarks primarily evaluate whether an agent can use
information from an interaction history when it becomes relevant, whereas
\textsc{VitaBench} evaluates whether an earlier decision about how that
information should persist was appropriate under its subsequently realized
future.

Table~\ref{tab:benchmark_comparison_extended} makes this distinction explicit.
The comparison concerns the released evaluation interfaces rather than whether
a benchmark contains phenomena such as updates, conflicts, or long-range
dependencies. Such phenomena can occur in several existing benchmarks without
providing counterfactual supervision over earlier representation choices.

\begin{table*}[t]
\caption{Comparison of memory-benchmark evaluation interfaces.
``Action scored'' means that an earlier memory-management decision is itself
an evaluation target. ``Same-state alternatives'' means that multiple candidate
actions are evaluated from the same pre-decision state. ``Shared-future CF''
means that these alternatives are evaluated under the same subsequent
interaction stream.}
\label{tab:benchmark_comparison_extended}
\centering
\footnotesize
\setlength{\tabcolsep}{5pt}
\begin{tabular}{lccccc}
\toprule
Benchmark
& \shortstack{Long-range\\use}
& \shortstack{Action\\scored}
& \shortstack{Same-state\\alternatives}
& \shortstack{Shared-future\\CF}
& \shortstack{Full-horizon\\value} \\
\midrule
LongMemEval
& yes & no & no & no & no \\
LoCoMo
& yes & no & no & no & no \\
MemoryAgentBench
& yes & no & no & no & no \\
MemOps
& yes & yes & no & no & no \\
\textsc{VitaBench}
& yes & yes & yes & yes & yes \\
\bottomrule
\end{tabular}
\end{table*}

LongMemEval, LoCoMo, and MemoryAgentBench evaluate complementary aspects of
long-term agent memory, including long-range retrieval, evidence reuse,
temporal reasoning, updates, conflicts, and incremental interaction
\citep{wu-2025-longmemeval,maharana-2024-locomo,
hu-2025-memoryagentbench}.
Their evaluation targets ultimately concern successful use of past information
in a later task. Even when an instance contains updates or conflicting
evidence, the released interface does not assign values to alternative
representation decisions that could have been made at an earlier time.
Consequently, a system can generally retain the available history and defer
representation choice until retrieval without being directly evaluated for
that earlier lifecycle decision.

MemOps also explicitly evaluates
memory operations such as creation, updating, deletion, and reflection
\citep{hao2026memopsbenchmarkinglifecyclememory}. It therefore moves beyond evaluating retrieval
alone and makes memory-management behavior itself an evaluation target.
The distinction from \textsc{VitaBench} is not whether lifecycle operations
exist, but how their consequences are evaluated.

At a \textsc{VitaBench} review point, the evaluator begins from one shared
pre-decision state, applies each candidate representation action independently,
and replays the same subsequent interaction stream for every branch. This
constructs a vector of action values
\[
    \{Q(s,a): a \in \mathcal{A}\}
\]
for the same state and realized future. The benchmark can therefore ask not
only whether a policy selected a plausible memory operation, but what would
have happened had it selected each alternative operation instead. Because the
future trajectory is shared across branches, differences in full-horizon
utility can be attributed to the representation decision being evaluated.

This counterfactual structure additionally permits quantities that are not
defined by operation labels alone. \emph{Lifecycle headroom} measures the
advantage of state-dependent representation choice over the best fixed action;
\emph{horizon reversal} measures whether the preferred action changes when
delayed consequences become visible; and policy regret measures the
full-horizon value lost by the action selected at a particular review state.
Thus, MemOps and \textsc{VitaBench} address related but complementary
questions: MemOps evaluates whether agents can perform appropriate memory
operations, while \textsc{VitaBench} evaluates the delayed consequences of
alternative representation decisions made before future use is known.

We do not claim that counterfactual lifecycle evaluation subsumes retrieval,
temporal reasoning, conflict resolution, or operation-level memory benchmarks.
Instead, \textsc{VitaBench} introduces an additional evaluation axis:
\emph{counterfactual evaluation of earlier representation choices under a
shared future trajectory}. Existing benchmarks remain useful for determining
whether an agent ultimately remembers and uses information correctly;
\textsc{VitaBench} is designed to determine whether the agent made appropriate
decisions about how that information should persist before its future use was
revealed.

\section{Robustness Across Diagnostic Regimes}
\label{app:sensitivity}

We evaluate the Diagnostic construction across 30 additional parameter
regimes varying exact-reuse delay, conflict-resolution probability,
recurrent-exact probability, and answer fidelity while retaining the same
60-thread streams, 120-episode horizon, action space, and balanced cost
accounting.

Across these regimes, lifecycle headroom averages $1.026 \pm 0.284$ and
horizon reversal averages $0.573 \pm 0.032$. Their weak correlation
(Pearson $r=0.015$) further indicates that the two diagnostics capture
complementary aspects of lifecycle decision structure.

\section{Reproducibility}
\label{app:repro}

All experiments in this paper are reproducible from the reported benchmark
configurations, data-generation procedures, random seeds, policy settings, and
evaluation protocols. Training, design, and held-out evaluation streams are
kept disjoint where applicable, and lifecycle policies use only information
available at the corresponding decision point.

\end{document}